\documentclass[11pt,a4paper]{article}
\pdfoutput=1
\usepackage{authblk}
\usepackage[hyphens]{url}
\usepackage[hyperref]{acl2017}
\usepackage{times}
\usepackage{latexsym}
\usepackage{amsmath,amsfonts,amssymb}
\usepackage{graphicx}
\usepackage{algorithm,algorithmic}
\usepackage{todonotes}

\aclfinalcopy 
\def\aclpaperid{741} 

\title{Watset: Automatic Induction of Synsets from a Graph of Synonyms}

\author[$\dag*$]{\bf Dmitry Ustalov}
\author[$\ddag$]{\bf Alexander Panchenko}
\author[$\ddag$]{\bf Chris Biemann}
\affil[$\dag$]{Institute of Natural Sciences and Mathematics, Ural Federal University, Russia}
\affil[$*$]{Krasovskii Institute of Mathematics and Mechanics, Russia}
\affil[$\ddag$]{Language Technology Group, Department of Informatics, Universit\"{a}t Hamburg, Germany}
\affil[ ]{\tt dmitry.ustalov@urfu.ru}
\affil[ ]{\tt \{panchenko,biemann\}@informatik.uni-hamburg.de}

\date{}

\usepackage{array}
\newcolumntype{R}[1]{>{\raggedleft\let\newline\\\arraybackslash\hspace{0pt}}m{#1}}

\def\watset#{\textsc{Watset}}
\def\cwtop#{CW\textsubscript{top}}
\def\cwlog#{CW\textsubscript{log}}
\def\cwnolog#{CW\textsubscript{nolog}}

\DeclareMathOperator{\ctx}{\mathrm{ctx}}
\DeclareMathOperator{\senses}{\text{senses}}

\newcommand\legend[1]{\fcolorbox{white}{#1}{\rule{0pt}{6pt}\rule{6pt}{0pt}}}
\definecolor{ggplotones}{HTML}{b3cde3}
\definecolor{ggplotcount}{HTML}{8c96c6}
\definecolor{ggplotsim}{HTML}{8856a7}

\begin{document}

\abovedisplayskip=4pt
\abovedisplayshortskip=1pt
\belowdisplayskip=1pt
\belowdisplayshortskip=1pt

\maketitle

\begin{abstract}
This paper presents a new graph-based approach that induces synsets using synonymy dictionaries and word embeddings. First, we build a weighted graph of synonyms extracted from commonly available resources, such as Wiktionary. Second, we apply word sense induction to deal with ambiguous words. Finally, we cluster the disambiguated version of the ambiguous input graph into synsets. Our meta-clustering approach lets us use an efficient hard clustering algorithm to perform a fuzzy clustering of the graph. Despite its simplicity, our approach shows excellent results, outperforming five competitive state-of-the-art methods in terms of F-score on three gold standard datasets for English and Russian derived from large-scale manually constructed lexical resources.
\end{abstract}

\section{Introduction}

A \textit{synset} is a set of mutual synonyms, which can be represented as a clique graph where nodes are words and edges are synonymy relations. Synsets represent word senses and are  building blocks of WordNet~\cite{Miller:95} and similar resources such as thesauri and lexical ontologies. These resources are crucial for many natural language processing applications that require common sense reasoning, such as information retrieval~\cite{Gong:05} and question answering~\cite{Kwok:01,Zhou:13}. However, for most languages, no manually-constructed resource is available that is comparable to the English WordNet in terms of coverage and quality. For instance, \newcite{Kiselev:15} present a comparative analysis of lexical resources available for the Russian language concluding that there is no resource compared to WordNet in terms of coverage and quality for Russian. This lack of linguistic resources for many languages urges the development of new methods for automatic construction of WordNet-like resources. The automatic methods foster construction and use of the new lexical resources.

Wikipedia\footnote{\url{http://www.wikipedia.org}}, Wiktionary\footnote{\url{http://www.wiktionary.org}}, OmegaWiki\footnote{\url{http://www.omegawiki.org}} and other collaboratively-created resources contain a large amount of lexical semantic information---yet designed to be human-readable and not formally structured. While semantic relations can be automatically extracted using tools such as DKPro JWKTL\footnote{\url{https://dkpro.github.io/dkpro-jwktl}} and Wikokit\footnote{\url{https://github.com/componavt/wikokit}}, words in these relations are not disambiguated. For instance, the synonymy pairs (\textit{bank}, \textit{streambank}) and (\textit{bank}, \textit{banking company}) will be connected via the word ``bank'', while they refer to the different senses. This problem stems from the fact that articles in Wiktionary and similar resources list undisambiguated synonyms. They are easy to disambiguate for humans while reading a dictionary article, but can be a source of errors for language processing systems.

The contribution of this paper is a novel approach that resolves ambiguities in the input graph to perform fuzzy clustering. The method takes as an input synonymy relations between potentially ambiguous terms available in human-readable dictionaries and transforms them into a machine readable representation in the form of disambiguated synsets. Our method, called \watset{}, is based on a new local-global meta-algorithm for fuzzy graph clustering. The underlying principle is to discover the word senses based on a \textit{local} graph clustering, and then to induce synsets using  \textit{global} sense clustering. We show that our method outperforms other methods for synset induction. The induced resource eliminates the need in manual synset construction and can be used to build WordNet-like semantic networks for under-resourced languages. An implementation of our method along with induced lexical resources is available online.\footnote{\url{https://github.com/dustalov/watset}}

\section{Related Work}\label{sec:related}

\textbf{Methods based on resource linking} surveyed by \newcite{Gurevych:16} gather various existing lexical resources and perform their linking to obtain a machine-readable repository of lexical semantic knowledge. For instance, BabelNet~\cite{Navigli:12:babelnet} relies in its core on a linking of WordNet and Wikipedia. UBY~\cite{Gurevych:12} is a general-purpose specification for the representation of lexical-semantic resources and links between them. The main advantage of our approach compared to the lexical resources is that no manual synset encoding is required. 

\textbf{Methods based on word sense induction} try to induce sense representations without the need for any initial lexical resource by extracting semantic relations from text. In particular, word sense induction (WSI) based on word ego networks clusters graphs of semantically related words~\cite{Lin:98,Pantel:02,Dorow:03,Veronis:04,Hope:13,Pelevina:16,Panchenko:17:eacl}, where each cluster corresponds to a word sense. An ego network consists of a single node (ego) together with the nodes they are connected to (alters) and all the edges among those alters~\cite{Everett:05}. In the case of WSI, such a network is a local neighborhood of one word. Nodes of the ego network are the words which are semantically similar to the target word.

\begin{figure*}[t]
  \centering
  \includegraphics[width=0.99\textwidth]{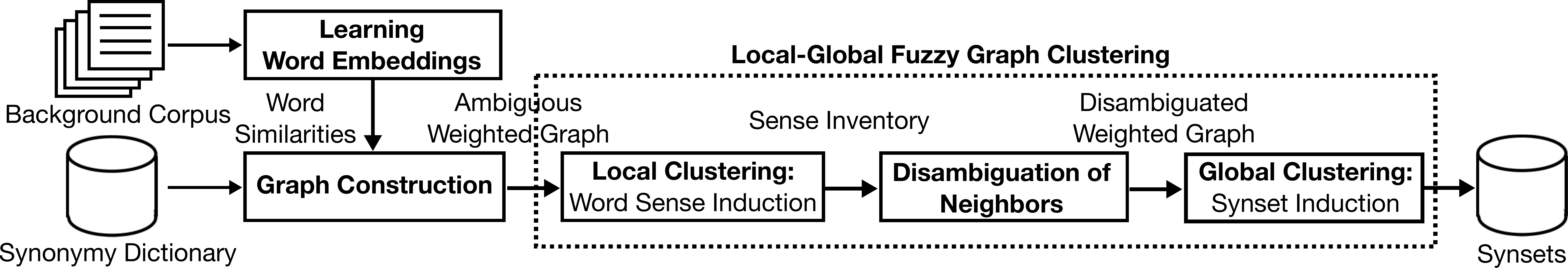}
  \caption{Outline of the \watset{} method for synset induction.}
  \label{fig:outline}
\end{figure*}

Such approaches are able to discover homonymous senses of words, e.g., ``bank'' as slope versus ``bank'' as organisation~\cite{DiMarco:12}. However, as the graphs are usually composed of semantically related words obtained using distributional methods~\cite{Baroni:10,Biemann:13:jobimtext}, the resulting clusters by no means can be considered synsets. Namely, (1) they contain words related not only via synonymy relation, but via a mixture of relations such as synonymy, hypernymy, co-hyponymy, antonymy, etc.~\cite{Heylen:08,Panchenko:11}; (2) clusters are not unique, i.e., one word can occur in clusters of different ego networks referring to the same sense, while in WordNet a word sense occurs only in a single synset.

In our synset induction method, we use word ego network clustering similarly as in word sense induction approaches, but apply them to a graph of semantically clean synonyms.

\textbf{Methods based on clustering of synonyms}, such as our approach, induce the resource from an ambiguous graph of synonyms where edges a extracted from manually-created resources. According to the best of our knowledge, most experiments either employed graph-based word sense induction applied to text-derived graphs or relied on a linking-based method that already assumes availability of a WordNet-like resource. A notable exception is the ECO approach by~\newcite{GoncaloOliveira:14}, which was applied to induce a WordNet of the Portuguese language called Onto.PT.\footnote{\url{http://ontopt.dei.uc.pt}} We compare to this approach and to five other state-of-the-art graph clustering algorithms as the baselines.

\textbf{ECO}~\cite{GoncaloOliveira:14} is a \textit{fuzzy} clustering algorithm that was used to induce synsets for a Portuguese WordNet from several available synonymy dictionaries. The algorithm starts by adding random noise to edge weights. Then, the approach applies Markov Clustering (see below) of this graph several times to estimate the probability of each word pair being in the same synset. Finally, candidate pairs over a certain threshold are added to output synsets.

\textbf{MaxMax}~\cite{Hope:13} is a \textit{fuzzy} clustering algorithm particularly designed for the word sense induction task. In a nutshell, pairs of nodes are grouped if they have a maximal mutual affinity. The algorithm starts by converting the undirected input graph into a directed graph by keeping the maximal affinity nodes of each node. Next, all nodes are marked as root nodes. Finally, for each root node, the following procedure is repeated: all transitive children of this root form a cluster and the root are marked as non-root nodes; a root node together with all its transitive children form a fuzzy cluster.

\textbf{Markov Clustering} (MCL)~\cite{vanDongen:00} is a \textit{hard} clustering algorithm for graphs based on simulation of stochastic flow in graphs. MCL simulates random walks within a graph by alternation of two operators called expansion and inflation, which recompute the class labels. Notably, it has been successfully used for the word sense induction task~\cite{Dorow:03}.

\textbf{Chinese Whispers} (CW)~\cite{Biemann:06} is a \textit{hard} clustering algorithm for weighted graphs that can be considered as a special case of MCL with a simplified class update step. At each iteration, the labels of all the nodes are updated according to the majority labels among the neighboring nodes. The algorithm has a meta-parameter that controls graph weights that can be set to three values: (1) \textit{top} sums over the neighborhood's classes; (2) \textit{nolog} downgrades the influence of a neighboring node by its degree or by (3) \textit{log} of its degree.

\textbf{Clique Percolation Method} (CPM)~\cite{Palla:05} is a \textit{fuzzy} clustering algorithm for unweighted graphs that builds up clusters from $k$-cliques corresponding to fully connected sub-graphs of $k$ nodes. While this method is only commonly used in social network analysis, we decided to add it to the comparison as synsets are essentially cliques of synonyms, which makes it natural to apply an algorithm based on clique detection.

\section{The \watset{} Method}

The goal of our method is to induce a set of unambiguous synsets by grouping individual ambiguous synonyms. An outline of the proposed approach is depicted in \figurename~\ref{fig:outline}. The method takes a dictionary of ambiguous synonymy relations and a text corpus as an input and outputs synsets. Note that the method can be used without a background corpus, yet as our experiments will show, corpus-based information improves the results when utilizing it for weighting the word graph's edges.

A synonymy dictionary can be perceived as a graph, where the nodes correspond to lexical entries (words) and the edges connect pairs of the nodes when the synonymy relation between them holds. The cliques in such a graph naturally form densely connected sets of synonyms corresponding to concepts~\cite{Gfeller:05}. Given the fact that solving the clique problem exactly in a graph is NP-complete~\cite{Bomze:99} and that these graphs typically contain tens of thousands of nodes, it is reasonable to use efficient hard graph clustering algorithms, like MCL and CW, for finding a global segmentation of the graph. However, the hard clustering property of these algorithm does not handle polysemy: while one word could have several senses, it will be assigned to only one cluster. To deal with this limitation, a word sense induction procedure is used to induce senses for all words, one at the time, to produce a disambiguated version of the graph where a word is now represented with one or many word senses. The concept of a disambiguated graph is described in \cite{Biemann:12}. Finally, the disambiguated word sense graph is clustered globally to induce synsets, which are hard clusters of word senses.

More specifically, the method consists of five steps  presented in \figurename~\ref{fig:outline}: (1) learning word embeddings; (2) constructing the ambiguous weighted graph of synonyms $G$; (3) inducing the word senses; (4) constructing the disambiguated weighted graph $G'$ by disambiguating of neighbors with respect to the induced word senses; (5) global clustering of the graph $G'$.

\subsection{Learning Word Embeddings}

Since different graph clustering algorithms are sensitive to edge weighting, we consider distributional semantic similarity based on word embeddings as a possible edge weighting approach for our synonymy graph. As we show further, this approach improves over unweighted versions and yields the best overall results.

\subsection{Construction of a Synonymy Graph}\label{sec:weights}

We construct the synonymy graph $G=(V, E)$ as follows. The set of nodes $V$ includes every lexeme appearing in the input synonymy dictionaries. The set of undirected edges $E$ is composed of all edges $(u, v) \in V \times V$ retrieved from one of the input synonymy dictionaries. We consider three edge weight representations:
\begin{itemize}
  \item \textbf{ones} that assigns every edge the constant weight of 1;
  \item \textbf{count} that weights the edge $(u, v)$ as the number of times the synonymy pair appeared in the input dictionaries;
  \item \textbf{sim} that assigns every edge $(u, v)$ a weight equal to the cosine similarity of skip-gram word vectors \cite{Mikolov:13}.
\end{itemize}

As the graph $G$ is likely to have polysemous words, the goal is to separate individual word senses using graph-based word sense induction.

\subsection{Local Clustering: Word Sense Induction}

\begin{figure}[t]
  \centering
  \includegraphics[width=.5\textwidth]{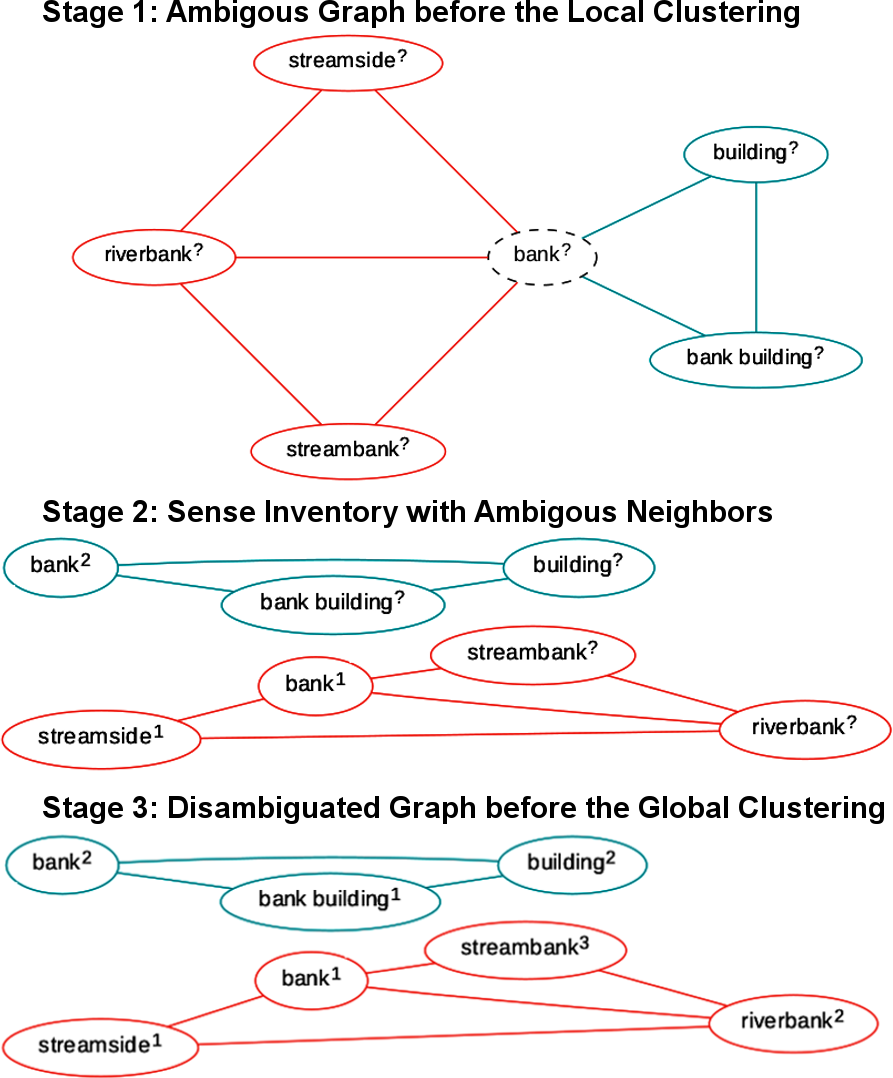}
  \vspace{-0.5em}
  \caption{Disambiguation of an ambiguous input graph using local clustering (WSI) to facilitate global clustering of words into synsets.}
  \label{fig:clustering}
\end{figure}

In order to facilitate global fuzzy clustering of the graph, we perform disambiguation of its ambiguous nodes as illustrated in \figurename~\ref{fig:clustering}. First, we use a graph-based word sense induction method that is similar to the curvature-based approach of~\newcite{Dorow:03}. In particular, removal of the nodes participating in many triangles tends to separate the original graph into several connected components. Thus, given a word $u$, we extract a network of its nearest neighbors from the synonymy graph $G$. Then, we remove the original word $u$ from this network and run a hard graph clustering algorithm that assigns one node to one and only one cluster. In our experiments, we test Chinese Whispers and Markov Clustering. The expected result of this is that each cluster represents a different sense of the word $u$, e.g.: 

{\begin{tabular}{lp{5cm}}
\textit{bank}\textsuperscript{1} & \{\textit{streambank}, \textit{riverbank}, \dots\}\\
\textit{bank}\textsuperscript{2} & \{\textit{bank company}, \dots\}\\
\textit{bank}\textsuperscript{3} & \{\textit{bank building}, \textit{building}, \dots\}\\
\textit{bank}\textsuperscript{4} & \{\textit{coin bank}, \textit{penny bank}, \dots\}
\end{tabular}}

We denote, e.g., bank\textsuperscript{1}, bank\textsuperscript{2} and other items as word senses referred to as $\senses(\text{bank})$. We denote as $\ctx(s)$ a cluster corresponding to the word sense $s$. Note that the context words have no sense labels. They are recovered by the disambiguation approach described next.

\subsection{Disambiguation of Neighbors}

Next, we disambiguate the neighbors of each induced sense. 
The previous step results in splitting word nodes into (one or more) sense nodes. However, nearest neighbors of each sense node are still ambiguous, e.g., (\textit{bank}\textsuperscript{3}, \textit{building}\textsuperscript{?}). To recover these sense labels of the neighboring words, we employ the following sense disambiguation approach proposed by \newcite{Faralli:16}. For each word $u$ in the context $\ctx(s)$ of the sense $s$, we find the most similar sense of that word $\hat{u}$ to the context. We use the cosine similarity measure between the context of the sense $s$ and the context of each candidate sense $u'$ of the word $u$:
\begin{equation*}
  \hat{u} =\!\! \underset{u' \in\, \senses(u)}{\arg\max}\!\! \cos(\ctx(s), \ctx(u'))\text{.}
\end{equation*}
A context $\ctx(\cdot)$ is represented by a sparse vector in a vector space of all ambiguous words of all contexts. The result is a disambiguated context $\widehat{\ctx}(s)$ in a space of \textit{disambiguated} words derived from its \textit{ambiguous} version $\ctx(s)$:
\begin{equation*}
  \widehat{\ctx}(s) = \{\hat{u} : u \in \ctx(s)\}\text{.}
\end{equation*} 

\subsection{Global Clustering: Synset Induction}

Finally, we construct the word sense graph $G'=(V', E')$ using the disambiguated senses instead of the original words and establishing the edges between these disambiguated senses: 
\begin{equation*}
V' = \bigcup_{u \in V}  \senses(u)\text{,}
\quad
E' = \bigcup_{s \in V'} \{s\} \times \widehat{\ctx}(s)\text{.}
\end{equation*}

Running a hard clustering algorithm on $G'$ produces the desired set of synsets as our final result. \figurename~\ref{fig:clustering} illustrates the process of disambiguation of an input ambiguous graph on the example of the word ``bank''. As one may observe, disambiguation of the nearest neighbors is a necessity to be able to construct a global version of the sense-aware graph. Note that current approaches to WSI, e.g.,~\cite{Veronis:04,Biemann:06,Hope:13}, do not perform this step, but perform only local clustering of the graph since they do not aim at a global representation of synsets.

\subsection{Local-Global Fuzzy Graph Clustering}

While we use our approach to synset induction in this work,   the core of our method is the ``local-global'' fuzzy graph clustering algorithm, which can be applied to arbitrary graphs (see \figurename~\ref{fig:outline}). This method, summarized in Algorithm~\ref{alg:watset}, takes an undirected graph $G=(V,E)$ as the input and outputs a set of fuzzy clusters of its nodes $V$. This is a meta-algorithm as it operates on top of two hard clustering algorithms denoted as $\text{Cluster}_\text{local}$ and $\text{Cluster}_\text{global}$, such as CW or MCL. At the first phase of the algorithm, for each node its senses are induced via ego network clustering (lines 1--7). Next, the disambiguation of each ego network is performed (lines 8--15). Finally, the fuzzy clusters are obtained by applying the hard clustering algorithm to the disambiguated graph (line 16). As a post-processing step, the sense labels can be removed to make the cluster elements subsets of $V$.

\begin{algorithm}[!ht]
\caption{\watset{} fuzzy graph clustering}
\label{alg:watset}
\begin{algorithmic}[1]
\REQUIRE{a set of nodes $V$ and a set of edges $E$.}
\ENSURE{a set of fuzzy clusters of $V$.} 
\FORALL{$u \in V$}
\STATE$C \gets \text{Cluster}_\text{local}(\text{Ego}(u))$  \COMMENT{$C = \{C_1,...\}$}
\FOR{$i \gets 1\dots|C|$}
\STATE$\ctx(u^i) \gets C_i$
\STATE$\senses(u) \gets \senses(u) \cup \{u^i\}$
\ENDFOR
\ENDFOR
\STATE$V' \gets \bigcup_{u \in V}  \senses(u)$
\FORALL{$s \in V'$}
\FORALL{$u \in \ctx(s)$}
\STATE$\hat{u} \gets\!\! \underset{u' \in\, \senses(u)}{\arg\max}\!\! \cos(\ctx(s), \ctx(u'))$
\ENDFOR
\STATE$\widehat{\ctx}(s) \gets \{\hat{u} : u \in \ctx(s)\}$
\ENDFOR
\STATE$E' \gets \bigcup_{s \in V'} \{s\} \times \widehat{\ctx}(s)$
\RETURN$\text{Cluster}_\text{global}(V', E')$
\end{algorithmic}
\end{algorithm}

\section{Evaluation}

We conduct our experiments on resources from two different languages. We evaluate our approach on two datasets for English to demonstrate its performance on a resource-rich language. Additionally, we evaluate it on two Russian datasets since Russian is a good example of an under-resourced language with a clear need for synset induction.

\subsection{Gold Standard Datasets}

For each language, we used two differently constructed lexical semantic resources listed in \tablename~\ref{tab:gold} to obtain gold standard synsets.

\paragraph{English.} We use \textbf{WordNet}\footnote{\url{https://wordnet.princeton.edu}}, a popular English lexical database constructed by expert lexicographers. WordNet contains general vocabulary and appears to be \textit{de facto} gold standard in similar tasks~\cite{Hope:13}. We used WordNet 3.1 to derive the synonymy pairs from synsets. Additionally, we use \textbf{BabelNet}\footnote{\url{http://www.babelnet.org}}, a large-scale multilingual semantic network constructed automatically using WordNet, Wikipedia and other resources. We retrieved all the synonymy pairs from the BabelNet 3.7 synsets marked as English. 

\paragraph{Russian.} As a lexical ontology for Russian, we use \textbf{RuWordNet}\footnote{\url{http://ruwordnet.ru/en}} \cite{Loukachevitch:16}, containing both general vocabulary and domain-specific synsets related to sport, finance, economics, etc. Up to a half of the words in this resource are multi-word expressions~\cite{Kiselev:15}, which is due to the coverage of domain-specific vocabulary. RuWordNet is a WordNet-like version of the RuThes thesaurus that is constructed in the traditional way, namely by a small group of expert lexicographers \cite{Loukachevitch:11}. In addition, we use \textbf{Yet Another RussNet}\footnote{\url{https://russianword.net/en}} (\textbf{YARN}) by \newcite{Braslavski:16} as another gold standard for Russian. The resource is constructed using crowdsourcing and mostly covers general vocabulary. Particularly, non-expert users are allowed to edit synsets in a collaborative way loosely supervised by a team of project curators. Due to the ongoing development of the resource, we selected as the gold standard only those synsets that were edited at least eight times in order to filter out noisy incomplete synsets.

\begin{table}[ht]
\resizebox{1.0\linewidth}{!}{
\centering
\begin{tabular}{lr|r|r|r}
\textbf{Resource} & & \textbf{\#~words} & \textbf{\#~synsets} & \textbf{\#~synonyms} \\\hline
WordNet   & En &     $148\,730$ &    $117\,659$ &     $152\,254$\\
BabelNet  & En & $11\,710\,137$ & $6\,667\,855$ & $28\,822\,400$\\
RuWordNet & Ru &     $110\,242$ &     $49\,492$ &     $278\,381$\\
YARN      & Ru &       $9\,141$ &      $2\,210$ &      $48\,291$\\
\end{tabular}
}
\caption{Statistics of the gold standard datasets.} 
\label{tab:gold}
\end{table}

\subsection{Evaluation Metrics}

To evaluate the quality of the induced synsets, we transformed them into binary synonymy relations and computed precision, recall, and F-score on the basis of the overlap of these binary relations with the binary relations from the gold standard datasets. Given a synset containing $n$ words, we generate a set of $\frac{n(n-1)}{2}$ pairs of synonyms. The F-score calculated this way is known as \textit{Paired F-score}~\cite{Manandhar:10,Hope:13}. The advantage of this measure compared to other cluster evaluation measures, such as \textit{Fuzzy B-Cubed}~\cite{Jurgens:13}, is its straightforward interpretability.


\subsection{Word Embeddings}

\paragraph{English.} We use the standard $300$-dimensional word embeddings trained on the $100$ billion tokens Google News corpus~\cite{Mikolov:13}.\footnote{\url{https://code.google.com/p/word2vec}}

\paragraph{Russian.} We use the $500$-dimensional word embeddings trained using the skip-gram model with negative sampling~\cite{Mikolov:13} using a context window size of $10$ with the minimal word frequency of $5$ on a $12.9$ billion tokens corpus of books. These embeddings were shown to produce state-of-the-art results in the RUSSE shared task\footnote{\url{http://www.dialog-21.ru/en/evaluation/2015/semantic_similarity}} and are part of the Russian Distributional Thesaurus (RDT) \cite{Panchenko:17:rdt}.\footnote{\url{http://russe.nlpub.ru/downloads}}

\subsection{Input Dictionary of Synonyms}

For each language, we constructed a synonymy graph using openly available language resources. The statistics of the graphs used as the input in the further experiments are shown in \tablename~\ref{tab:raw}.

\paragraph{English.} Synonyms were extracted from the English Wiktionary\footnote{We used the Wiktionary dumps of February 1, 2017.}, which is the largest Wiktionary at the present moment in terms of the lexical coverage, using the DKPro~JWKTL tool by~\newcite{Zesch:08}. English words have been extracted from the dump.

\paragraph{Russian.} Synonyms from three sources were combined to improve lexical coverage of the input dictionary and to enforce confidence in jointly observed synonyms: (1) synonyms listed in the Russian Wiktionary extracted using the Wikokit tool by~\newcite{Krizhanovsky:13}; (2) the dictionary of~\newcite{Abramov:99}; and (3) the Universal Dictionary of Concepts~\cite{Dikonov:13}. While the two latter resources are specific to Russian, Wiktionary is available for most languages. Note that the same input synonymy dictionaries were used by authors of YARN to construct synsets using crowdsourcing. The results on the YARN dataset show how close an automatic synset induction method can approximate manually created synsets provided the same starting material.\footnote{We used the YARN dumps of February 7, 2017.}

\begin{table}[ht]
\centering
\begin{tabular}{l|r|r}
\textbf{Language} & \textbf{\#~words} & \textbf{\#~synonyms} \\\hline
English & $243\,840$ & $212\,163$\\
Russian &  $83\,092$ & $211\,986$\\
\end{tabular}
\caption{Statistics of the input datasets.} 
\label{tab:raw}
\end{table}

\section{Results}

\begin{figure*}[!ht]
  \centering
  \includegraphics[scale=.8]{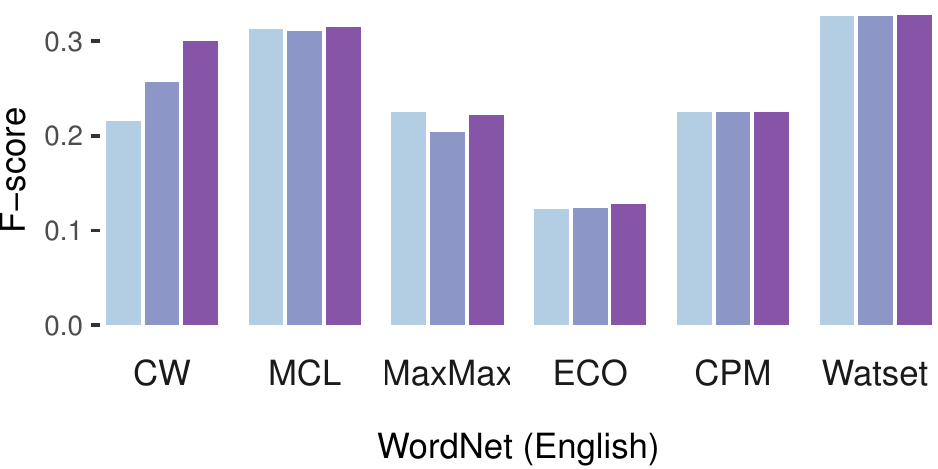}  
  \quad
  \includegraphics[scale=.8]{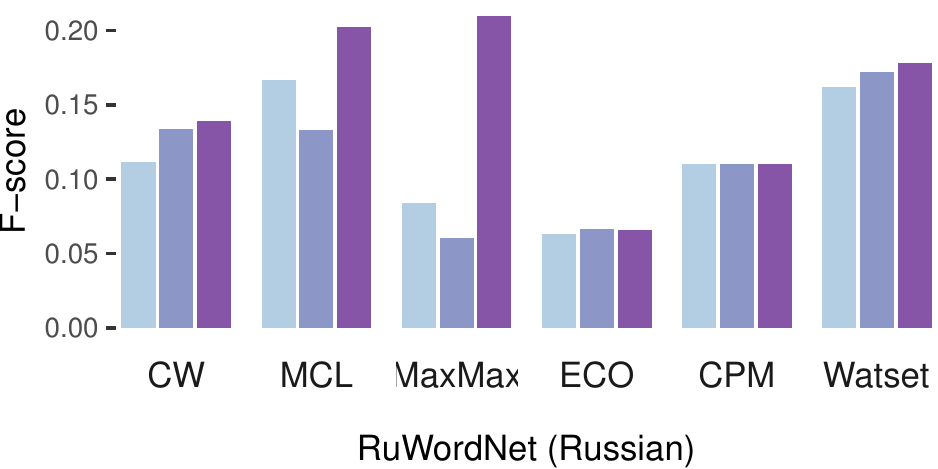}      
  \\\medskip
  \includegraphics[scale=.8]{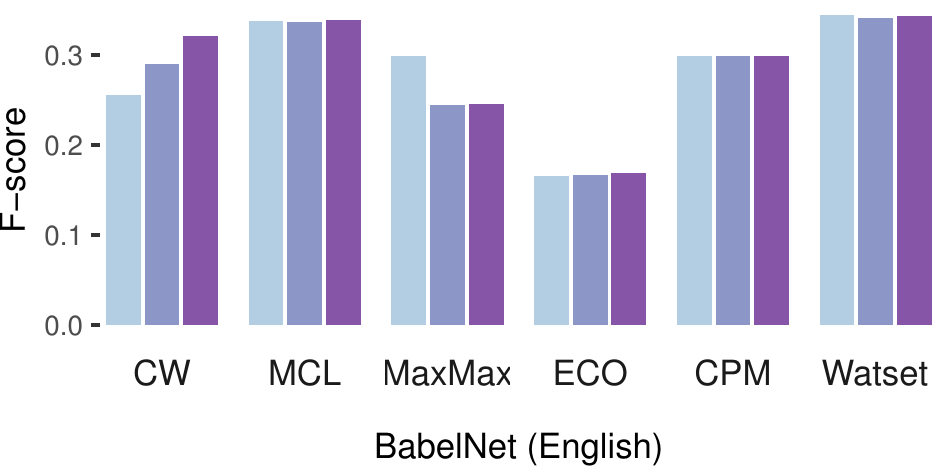} 
  \quad
  \includegraphics[scale=.8]{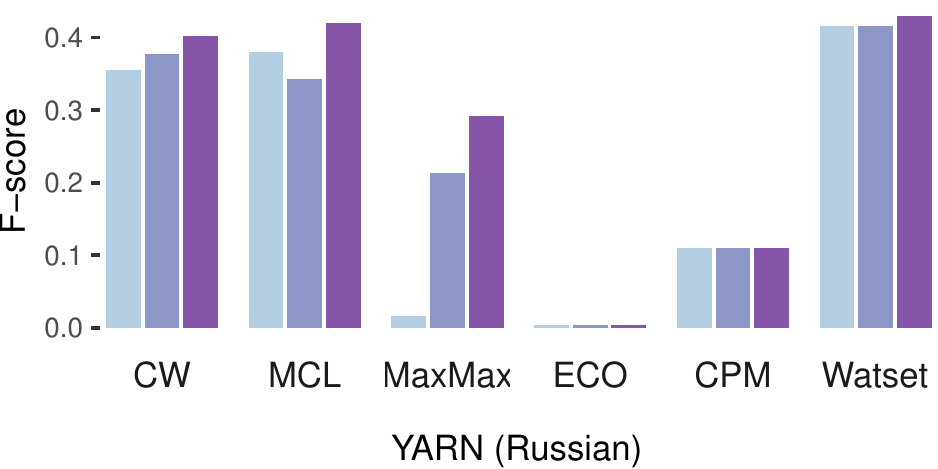}     
  \caption{Impact of the different graph weighting schemas on the performance of synset induction: \legend{ggplotones}\,ones, \legend{ggplotcount}\,count, \legend{ggplotsim}\,sim. Each bar corresponds to the top performance of a method in Tables~\ref{tab:english} and~\ref{tab:russian}.} 
  \label{fig:weights}
\end{figure*}

We compare \watset{} with five state-of-the art graph clustering methods presented in Section~\ref{sec:related}: Chinese Whispers (CW), Markov Clustering (MCL), MaxMax, ECO clustering, and the clique percolation method (CPM). The first two algorithms perform hard clustering, while the last three are fuzzy clustering methods just like our method. While the hard clustering algorithms are able to discover clusters which correspond to synsets composed of unambigous words, they can produce wrong results in the presence of lexical ambiguity (one node belongs to several synsets). In our experiments, we rely on our own implementation of MaxMax and ECO as reference implementations are not available. For CW\footnote{\url{https://www.github.com/uhh-lt/chinese-whispers}}, MCL\footnote{\url{http://java-ml.sourceforge.net}} and CPM\footnote{\url{https://networkx.github.io}}, available implementations have been used. During the evaluation, we delete clusters equal or larger than the threshold of 150 words as they hardly can represent any meaningful synset. The notation \watset{[MCL, \cwtop{}]} means using MCL for local clustering and Chinese Whispers in the \textit{top} mode for global clustering.


\subsection{Impact of Graph Weighting Schema}

\figurename~\ref{fig:weights} presents an overview of the evaluation results on both datasets. The first step, common for all of the tested synset induction methods, is graph construction. Thus, we started with an analysis of three ways to weight edges of the graph introduced in Section~\ref{sec:weights}: binary scores (\textit{ones}), frequencies (\textit{count}), and semantic similarity scores (\textit{sim}) based on word vector similarity. Results across various configurations and methods indicate that using the weights based on the similarity scores provided by word embeddings is the best strategy for all methods except MaxMax on the English datasets. However, its performance using the \textit{ones} weighting does not exceed the other methods using the \textit{sim} weighting. Therefore, we report all further results on the basis of the \textit{sim} weights. The edge weighting scheme impacts Russian more for most algorithms. The CW algorithm however remains sensitive to the weighting also for the English dataset due to its randomized nature.

\begin{table*}[!ht]
\centering
\resizebox{1.0\linewidth}{!}{
\begin{tabular}{l*{3}{|r}*{6}{|c}}
\multicolumn{4}{c|}{} & \multicolumn{3}{c|}{\textbf{WordNet}} & \multicolumn{3}{c}{\textbf{BabelNet}}\\
\cline{5-10}
\textbf{Method} & \textbf{\#~words} & \textbf{\#~synsets} & \textbf{\#~synonyms} & \textbf{P} & \textbf{R} & \textbf{F1} & \textbf{P} & \textbf{R} & \textbf{F1}\\\hline
\watset{[MCL, MCL]} & $243\,840$ & $112\,267$ & $345\,883$ &
$0.345$ & $\mathbf{0.308}$ & $\mathbf{0.325}$ & $0.400$ & $0.301$ & $\underline{\mathbf{0.343}}$ \\
MCL & $243\,840$ & $84\,679$ & $387\,315$ &
$0.342$ & $0.291$ & $0.314$ & $0.390$ & $0.300$ & $\mathbf{0.339}$ \\
\watset{[MCL, \cwlog{}]} & $243\,840$ & $105\,631$ & $431\,085$ &
$0.314$ & $\underline{\mathbf{0.325}}$ & $\mathbf{0.319}$ & $0.359$ & $\mathbf{0.312}$ & $\mathbf{0.334}$ \\
\cwtop{} & $243\,840$ & $77\,879$ & $539\,753$ &
$0.285$ & $\mathbf{0.317}$ & $0.300$ & $0.326$ & $\underline{\mathbf{0.317}}$ & $0.321$ \\
\watset{[\cwlog{}, MCL]} & $243\,840$ & $164\,689$ & $227\,906$ &
$\mathbf{0.394}$ & $0.280$ & $\underline{\mathbf{0.327}}$ & $\mathbf{0.439}$ & $0.245$ & $0.314$ \\
\watset{[\cwlog{}, \cwlog{}]} & $243\,840$ & $164\,667$ & $228\,523$ &
$0.392$ & $0.280$ & $\underline{\mathbf{0.327}}$ & $\mathbf{0.439}$ & $0.245$ & $0.314$ \\
CPM{\textsubscript{$k\!=\!2$}} & $186\,896$ & $67\,109$ & $317\,293$ &
$\mathbf{0.561}$ & $0.141$ & $0.225$ & $\mathbf{0.492}$ & $0.214$ & $0.299$ \\
MaxMax & $219\,892$ & $73\,929$ & $797\,743$ &
$0.176$ & $0.300$ & $0.222$ & $0.202$ & $\mathbf{0.313}$ & $0.245$ \\
ECO & $243\,840$ & $171\,773$ & $84\,372$ &
$\underline{\mathbf{0.784}}$ & $0.069$ & $0.128$ & $\underline{\mathbf{0.699}}$ & $0.096$ & $0.169$ \\
\end{tabular}
}
\caption{Comparison of the synset induction methods on datasets for English. All methods rely on the similarity edge weighting (\textit{sim}); best configurations of each method in terms of F-scores are shown for each dataset. Results are sorted by F-score on BabelNet, top three values of each metric are boldfaced.}
\label{tab:english}
\end{table*}
\begin{table*}[!ht]
\centering
\resizebox{1.0\linewidth}{!}{
\begin{tabular}{l*{3}{|r}*{6}{|c}}
\multicolumn{4}{c|}{} & \multicolumn{3}{c|}{\textbf{RuWordNet}} & \multicolumn{3}{c}{\textbf{YARN}}\\
\cline{5-10}
\textbf{Method} & \textbf{\#~words} & \textbf{\#~synsets} & \textbf{\#~synonyms} & \textbf{P} & \textbf{R} & \textbf{F1} & \textbf{P} & \textbf{R} & \textbf{F1}\\\hline
\watset{[\cwnolog{}, MCL]} & $83\,092$ & $55\,369$ & $332\,727$ &
$0.120$ & $\mathbf{0.349}$ & $\mathbf{0.178}$ & $0.402$ & $\mathbf{0.463}$ & $\underline{\mathbf{0.430}}$ \\
\watset{[MCL, MCL]} & $83\,092$ & $36\,217$ & $403\,068$ &
$0.111$ & $0.341$ & $0.168$ & $0.405$ & $0.455$ & $\mathbf{0.428}$ \\
\watset{[\cwtop{}, \cwlog{}]} & $83\,092$ & $55\,319$ & $341\,043$ &
$0.116$ & $\mathbf{0.351}$ & $0.174$ & $0.386$ & $\mathbf{0.474}$ & $\mathbf{0.425}$ \\
MCL & $83\,092$ & $21\,973$ & $353\,848$ &
$0.155$ & $0.291$ & $\mathbf{0.203}$ & $0.550$ & $0.340$ & $0.420$ \\
\watset{[MCL, \cwtop{}]} & $83\,092$ & $34\,702$ & $473\,135$ &
$0.097$ & $\underline{\mathbf{0.361}}$ & $0.153$ & $0.351$ & $\underline{\mathbf{0.496}}$ & $0.411$ \\
\cwnolog{} & $83\,092$ & $19\,124$ & $672\,076$ &
$0.087$ & $0.342$ & $0.139$ & $0.364$ & $0.451$ & $0.403$ \\
MaxMax & $83\,092$ & $27\,011$ & $461\,748$ &
$\mathbf{0.176}$ & $0.261$ & $\underline{\mathbf{0.210}}$ & $\mathbf{0.582}$ & $0.195$ & $0.292$ \\
CPM{\textsubscript{$k\!=\!3$}} & $15\,555$ & $4\,000$ & $45\,231$ &
$\mathbf{0.234}$ & $0.072$ & $0.111$ & $\mathbf{0.626}$ & $0.060$ & $0.110$ \\
ECO & $83\,092$ & $67\,645$ & $18\,362$ &
$\underline{\mathbf{0.724}}$ & $0.034$ & $0.066$ & $\underline{\mathbf{0.904}}$ & $0.002$ & $0.004$ \\
\end{tabular}
}
\caption{Results on Russian sorted by F-score on YARN, top three values of each metric are boldfaced.}
\label{tab:russian}
\end{table*}

\subsection{Comparative Analysis}

\tablename~\ref{tab:english} and~\ref{tab:russian} present evaluation results for both languages. For each method, we show the best configurations in terms of F-score. One may note that the granularity of the resulting synsets, especially for Russian, is very different, ranging from 4\,000 synsets for the CPM{\textsubscript{$k\!=\!3$}} method to 67\,645 induced by the ECO method. Both tables report the number of words, synsets and synonyms after pruning huge clusters larger than 150 words. Without this pruning, the MaxMax and CPM methods tend to discover giant components obtaining almost zero precision as we generate all possible pairs of nodes in such clusters. The other methods did not show such behavior.

\watset{} robustly outperforms all other methods according to F-score on both English datasets (\tablename~\ref{tab:english}) and on the YARN dataset for Russian (\tablename~\ref{tab:russian}). Also, it outperforms all other methods according to recall on both Russian datasets. The disambiguation of the input graph performed by the \watset{} method splits nodes belonging to several local communities to several nodes, significantly facilitating the clustering task otherwise complicated by the presence of the hubs that wrongly link semantically unrelated nodes.

Interestingly, in all the cases, the toughest competitor was a hard clustering algorithm---MCL \cite{vanDongen:00}. We observed that the ``plain'' MCL successfully groups monosemous words, but isolates the neighborhood of polysemous words, which results in the recall drop in comparison to \watset{}. CW operates faster due to a simplified update step. On the same graph, CW tends to produce larger clusters than MCL. This leads to a higher recall of ``plain'' CW as compared to the ``plain'' MCL, at the cost of lower precision.

Using MCL instead of CW for sense induction in \watset{} expectedly produces more fine-grained senses. However, at the global clustering step, these senses erroneously tend to form coarse-grained synsets connecting unrelated senses of the ambiguous words. This explains the generally higher recall of \watset{[MCL, $\cdot$]}. Despite the randomized nature of CW, variance across runs do not affect the overall ranking: The rank of different versions of CW (\textit{log, nolog, top}) can change, while the rank of the best CW configuration compared to other methods remains the same.

The MaxMax algorithm shows mixed results. On the one hand, it outputs large clusters uniting more than hundred nodes. This inevitably leads to a high recall, as it is clearly seen in the results for Russian because such synsets still pass under our cluster size threshold of 150 words. Its synsets on English datasets are even larger and get pruned, which results in low recall. On the other hand, smaller synsets having at most 10--15 words were identified correctly. MaxMax appears to be extremely sensible to edge weighting, which also complicates its practical use.

The CPM algorithm showed unsatisfactory results, emitting giant components encompassing thousands of words. Such clusters were automatically pruned, but the remaining clusters are relatively correctly built synsets, which is confirmed by the high values of precision. When increasing the minimal number of elements in the clique $k$, recall improves, but at the cost of a dramatic precision drop. We suppose that the network structure assumptions exploited by CPM do not accurately model the structure of our synonymy graphs.

Finally, the ECO method yielded the worst results because the most cluster candidates failed to pass through the constant threshold used for estimating whether a pair of words should be included in the same cluster. Most synsets produced by this method were trivial, i.e., containing only a single word. The remaining synsets for both languages have at most three words that have been connected by a chance due to the edge noising procedure used in this method resulting in low recall.

\section{Discussion}

\paragraph{On the absolute scores.} The results obtained on all gold standards (\figurename~\ref{fig:weights}) show similar trends in terms of relative ranking of the methods. Yet absolute scores of YARN and RuWordNet are substantially different due to the inherent difference of these datasets. RuWordNet is more domain-specific in terms of vocabulary, so our input set of generic synonymy dictionaries has a limited coverage on this dataset. On the other hand, recall calculated on YARN is substantially higher as this resource was manually built on the basis of synonymy dictionaries used in our experiments.

The reason for low absolute numbers in evaluations is due to an inherent vocabulary mismatch between the input dictionaries of synonyms and the gold datasets. To validate this hypothesis, we performed a cross-resource evaluation presented in \tablename~\ref{tab:xres}. The low performance of the cross-evaluation of the two resources supports the hypothesis: no single resource for Russian can obtain high recall scores on another one. Surprisingly, even BabelNet, which integrates most of available lexical resources, still does not reach a recall substantially larger than 0.5.\footnote{We used BabelNet 3.7 extracting  all 3\,497\,327 synsets that were marked as Russian.} Note that the results of this cross-dataset evaluation are not directly comparable to results in \tablename~\ref{tab:russian} since in our experiments we use much smaller input dictionaries than those used by BabelNet.

\begin{table}[t]
\resizebox{\linewidth}{!}{
\centering
\begin{tabular}{ll*{3}{|c}}
\textbf{Resource}   & & \textbf{P} & \textbf{R} & \textbf{F1} \\\hline
BabelNet on WordNet    & En & $0.729$ & $0.998$ & $0.843$ \\
WordNet on BabelNet    & En & $0.998$ & $0.699$ & $0.822$ \\ \hline
YARN on RuWordNet      & Ru & $0.164$ & $0.162$ & $0.163$ \\
BabelNet on RuWordNet  & Ru & $0.348$ & $0.409$ & $0.376$ \\\hline
RuWordNet on YARN      & Ru & $0.670$ & $0.121$ & $0.205$ \\
BabelNet on YARN       & Ru & $0.515$ & $0.109$ & $0.180$ \\
\end{tabular}
}
\caption{Performance of lexical resources cross-evaluated against each other.}
\label{tab:xres}
\end{table}

\paragraph{On sparseness of the input dictionary.} \tablename~\ref{tab:sample} presents some examples of the obtained synsets of various sizes for the top \watset{} configuration on both languages. As one might observe, the quality of the results is highly plausible. However, one limitation of all approaches considered in this paper is the dependence on the completeness of the input dictionary of synonyms. In some parts of the input synonymy graph, important bridges between words can be missing, leading to smaller-than-desired synsets. A promising extension of the present methodology is using distributional models to enhance connectivity of the graph by cautiously adding extra relations.

\begin{table}[ht]
\resizebox{\linewidth}{!}{
\centering
\small
\begin{tabular}{c|p{60mm}}
\textbf{Size} & \textbf{Synset}\\\hline
2 & \{\textit{decimal point}, \textit{dot}\}\\
3 & \{\textit{gullet}, \textit{throat}, \textit{food pipe}\}\\
4 & \{\textit{microwave meal}, \textit{ready meal}, \textit{TV dinner}, \textit{frozen dinner}\}\\
5 & \{\textit{objective case}, \textit{accusative case}, \textit{oblique case}, \textit{object case}, \textit{accusative}\}\\
6 & \{\textit{radio theater}, \textit{dramatized audiobook}, \textit{audio theater}, \textit{radio play}, \textit{radio drama}, \textit{audio play}\}\\
\end{tabular}
}
\caption{Sample synsets induced by the \watset{[MCL, MCL]} method for English.}
\label{tab:sample}
\end{table}

\section{Conclusion}

We presented a new robust approach to fuzzy graph clustering that relies on hard graph clustering. Using ego network clustering, the nodes belonging to several local communities are split into several nodes each belonging to one community. The transformed ``disambiguated'' graph is then clustered using an efficient hard graph clustering algorithm, obtaining a fuzzy clustering as the result. The disambiguated graph facilitates clustering as it contains fewer hubs connecting unrelated nodes from different communities. We apply this meta clustering algorithm to the task of synset induction on two languages, obtaining the best results on three datasets and competitive results on one dataset in terms of F-score as compared to five state-of-the-art graph clustering methods.

\section*{Acknowledgments}

We acknowledge the support of the Deutsche Forschungsgemeinschaft (DFG) foundation under the ``JOIN-T'' project, the DAAD, the RFBR under the project no.~16-37-00354 mol\_a, and the RFH under the project no.~16-04-12019. We also thank three anonymous reviewers for their helpful comments, Andrew Krizhanovsky for providing a parsed Wiktionary, Natalia Loukachevitch for the provided RuWordNet dataset, and Denis Shirgin who suggested the \watset{} name.

\bibliography{watset.acl2017}
\bibliographystyle{acl_natbib}

\end{document}